\documentclass[runningheads]{llncs}
\usepackage{graphicx}
\usepackage{graphbox}
\usepackage{amssymb}
\usepackage{booktabs}
\usepackage{multirow}
\usepackage{siunitx}
\usepackage[draft,inline,nomargin,index]{fixme}
\fxsetup{theme=color, mode=multiuser}
\FXRegisterAuthor{f}{felix}{\color{red}fe}
\FXRegisterAuthor{t}{tobias}{\color{blue}to}
\FXRegisterAuthor{o}{olde}{\color{teal}ol}
\usepackage[shortcuts, acronym, toc, nonumberlist=false]{glossaries}
\makenoidxglossaries
 \newacronym
 {dnn}{DNN}{deep neural network} 
 \newacronym
 {ml}{ML}{machine learning}
 \newacronym
 {dl}{DL}{deep learning}
 \newacronym
 {rbf}{RBF}{radial basis function}
 
  \newacronym
 {bim}{BIM}{basic iterative method}
 \newacronym
 {cw}{CW}{Carlini-Wagner}
  \newacronym
 {fgsm}{FGSM}{fast gradient sign method}
  \newacronym
 {svm}{SVM}{support vector machine}
\begin{document}
\title{Learning to Detect Adversarial Examples Based on Class Scores}
\author{Tobias Uelwer\inst{1} \and
Felix Michels\inst{1} \and
Oliver De Candido\inst{2} \orcidID{0000-0002-9523-7777}
}

\authorrunning{T. Uelwer et al.}

\institute{Department of Computer Science, Heinrich Heine University D\"usseldorf, Germany
\email{\{tobias.uelwer, felix.michels\}@hhu.de}\\
\and
Department of Electrical and Computer Engineering, Technical~University~of~Munich, Germany\\
\email{oliver.de-candido@tum.de}}
\maketitle              % typeset the header of the contribution
\begin{abstract}
Given the increasing threat of adversarial attacks on \glspl{dnn}, research on efficient detection methods is more important than ever.
In this work, we take a closer look at adversarial attack detection based on the class scores of an already trained classification model.
We propose to train a \gls{svm} on the class scores to detect adversarial examples.
Our method is able to detect adversarial examples generated by various attacks, and can be easily adopted to a plethora of deep classification models.
We show that our approach yields an improved detection rate compared to an existing method, whilst being easy to implement.
We perform an extensive empirical analysis on different deep classification models, investigating various state-of-the-art adversarial attacks.
Moreover, we observe that our proposed method is better at detecting a combination of adversarial attacks.
This work indicates the potential of detecting various adversarial attacks simply by using the class scores of an already trained classification model.

\end{abstract}

%  Reset the gls definitions
\glsresetall

\section{Introduction}
In recent years, \glspl{dnn} have demonstrated exemplary performance on image classification tasks~\cite{simonyan2014very,szegedy2015going,he2016deep}.
However, in their seminal work, Goodfellow et al.~\cite{goodfellow2014explaining} have shown that these deep models can easily be deceived into misclassifying inputs by using adversarial examples (i.e., perturbed input images). 
These perturbed images can be crafted in a white-box setting where the attacker is given access to the (unnormalized) class scores of the \gls{dnn} classifier, and is able to calculate gradients with respect to the input image.
In contrast, in the black-box setting, the attacker is only able to observe the models final decision and cannot perform any gradient calculations.
In this work, we consider the white-box setting as well as the black-box setting.

The importance of adversarial defenses is evident, if we consider the wide variety of safety-critical domains where \glspl{dnn} are employed, e.g., in medical imaging~\cite{paschali2018generalizability}, in autonomous driving~\cite{sitawarin2018darts,eykholt2018robust}, or in face recognition~\cite{dong2019efficient}.
As defined in~\cite{xu2020adversarial}, there are three main adversarial defense mechanisms: (i) gradient masking, (ii) robust optimization, and (iii) adversarial example detection.
In this paper, we focus on the latter mechanism, and propose a method to detect adversarial examples based on the output class scores.

\subsection{Related Work}
One of the first methods to robustify the optimization of \glspl{dnn} was proposed by Goodfellow et al. in \cite{goodfellow2014explaining}.  
The main idea is to adopt adversarial training by augmenting the training dataset with adversarial examples.
This robustifies the  model and it has been shown that adversarial training makes the model less susceptible to adversarial attacks of the same type.
Grosse et al. \cite{grosse2017statistical} suggest adversarial retraining with an additional class for the adversarial examples.
These retraining methods, however, are very time consuming as one not only has to generate many adversarial examples to train on, but one also needs to retrain the whole \gls{dnn}.

To overcome the computational costs required to robustify optimization via adversarial training, researchers have focused on adversarial example detection instead.
On the one hand, one can attempt to detect the adversarial examples in the input domain.
Thereby, training an additional model to classify whether the input image is clean or not \cite{metzen2017detecting}.
Another possibility is to detect the adversarial examples in the input space using a smaller \gls{dnn} classifier, which is easier to train than the original \gls{dnn}.

On the other hand, one can investigate the latent representations within the \gls{dnn} to detect adversarial examples.
Papernot and McDaniel~\cite{papernot2018deep} propose training a $k$-nearest neighbors classifier on the representations in each layer of a \gls{dnn} to estimate whether a given input conforms with the labels of the surrounding representations.
They show that this method provides a better estimate for the confidence of the \gls{dnn}, and can also detect adversarial examples.
In the limit, one can use the output class scores to detect adversarial examples.
Kwon et al.~\cite{kwon2021classification} propose a method to detect adversarial examples by thresholding the difference of the highest and second highest class scores at the output of the \gls{dnn}.

\subsection{Contributions}
The contributions of this work can be summarized as follows:
\begin{enumerate}
    \item We propose to train a \gls{svm}~\cite{cortes1995support} with a \gls{rbf} kernel on the class scores to detect adversarial examples.
    \item We empirically evaluate our method against the adversarial detection method by Kwon et al. \cite{kwon2021classification}, and show that our method improves the detection performance of different adversarial attacks on various deep classification models.
    \item Furthermore, we observe that our method is better at detecting a combination of two adversarial attacks, compared with \cite{kwon2021classification}.
\end{enumerate}

\section{Detecting Adversarial Examples From Class Scores}\label{sec:method}
For an input image $x$ we denote the class scores that the attacked classification model $F$ predicts by $F(x)$.
Instead of retraining the whole model from scratch, we take advantage of the fact that adversarial attacks can be detected by classifying the class scores predicted by $F$.
To this end, we train an \gls{svm} \cite{cortes1995support} with an \gls{rbf} kernel on the class scores to detect whether unseen images have been altered or not.
In order to perform adversarial example detection we proceed as follows:
\begin{enumerate}
\item We construct a dataset $X_\text{adv}= \{\tilde x_1,\dots, \tilde  x_n\}$ that contains a perturbed instance of each image in the training dataset $X_\text{train}=\{x_1,\dots, x_n\}$.
\item We train the \gls{svm} on $X_\text{scores}=\{F(x_1),\dots, F(x_n), F(\tilde x_1),\dots, F(\tilde x_n)\}$ to predict the label $1$ for samples coming from $X_\text{train}$ and $-1$ for samples from $X_\text{adv}$ based on the outputs of $F$.
\item At test time, the trained \gls{svm} can then be used to detect adversarial examples based on the class score $F({x}_\text{new})$ where $x_\text{new}$ is a new (possibly adversarial) input image.
\end{enumerate}
To the best of our knowledge, only Kwon et al.~\cite{kwon2021classification} also use class scores to classify adversarial and non-adversarial inputs.
In contrast to our method, they solely threshold the difference between the two largest scores.
We argue that this approach discards valuable information contained in the class scores.
Moreover, they choose the threshold value by hand, instead of optimizing for it.
Our approach can be seen as a generalization to this method as we classify adversarial and non-adversarial input images by taking all of the class scores into account.

\section{Experimental Setup}
We calculate adversarial examples on three different pre-trained \gls{dnn} architectures. 
Namely, the VGG19~\cite{simonyan2014very}, GoogLeNet~\cite{szegedy2015going}, and ResNet18~\cite{he2016deep}.
All models were trained on the CIFAR-10 image recognition dataset \cite{krizhevsky2009learning}.
To evaluate the efficiency of our proposed detection method, we consider adversarial examples produced by the \gls{bim}~\cite{kurakin2016adversarial}, the \gls{fgsm}~\cite{goodfellow2014explaining}, the boundary attack \cite{brendel2018decisionbased}, and the \gls{cw} method~\cite{carlini2017towards}. 
\gls{bim}, \gls{fgsm} and \gls{cw} are white-box attacks, whereas the boundary attack is a black-box attack.
We use the implementations provided by the Python~3 package Foolbox~\cite{rauber2017foolbox,rauber2020foolbox}.

Moreover, we analyze the detection performance on a combination of two different attacks at the same time, i.e., we also consider detecting adversarial perturbations produced by \gls{cw} and \gls{bim}, \gls{cw} and \gls{fgsm}, boundary attack and \gls{bim}, and boundary attack and \gls{fgsm}.
When detecting a single attack, the training dataset consists of original images and adversarial images in the ratio $1:1$, whereas when detecting two attacks the ratio is $2:1:1$ for original images, adversarial images from the first attack and adversarial images from the second attack, respectively.

For \gls{fgsm}, we choose the step size $\epsilon$ individually for each model in a way that the attack is successful in 50\% of the cases.
We run \gls{bim} for $100$ steps with a relative step size of $0.2$ while allowing random starts. 
Again, we tune the maximum Euclidean norm of the perturbation for each model to achieve a success rate of approximately 95\%.
For the \gls{cw} method we tune the step size and the number of steps for each model until all perturbed inputs were misclassified.
We run the boundary attack for $\num{25000}$ steps which is the default value of the parameter in the Foolbox implementation. 

To determine the threshold used in the detection method from Kwon et al.~\cite{kwon2021classification}, we use a decision stump trained using the Gini index \cite{friedman2001greedy}.
The regularization parameter $C$ of the \gls{svm} and the kernel-bandwidth $\gamma$ of the \gls{rbf} kernel are optimized using a coarse-to-fine gridsearch.

\section{Results}
On the original test dataset (in absence of adversarial examples), the VGG19 achieves a classification accuracy of $93.95\%$, the GoogLeNet $92.85\%$, and the ResNet18 $93.07\%$. 
In Table~\ref{tab:acc}, we summarize the performance of the adversarial attacks, i.e., the classification accuracy on the test dataset after the attack.
It also shows the average Euclidean norm of the perturbations for each classification model with the classification performance in the unattacked setting.
In our experiments, the \gls{cw} method and the boundary method produced images with the lowest average perturbation norm across all pre-trained \gls{dnn} classifiers.
Moreover, the \gls{cw} method is more consistent at lowering the classification accuracy and strictly lower than the other attacks for each \gls{dnn}.
These advantages, however, come at the price of higher computational effort.

\begin{table}[]
    \caption{Comparison of the classification accuracies on the adversarial examples for each model and the average Euclidean norm of the added perturbations.}
    \centering
    \addtolength{\tabcolsep}{4pt}
    \begin{tabular}{lrrrrrr} 
    \toprule
     & \multicolumn{3}{c}{Accuracy on adversarial examples}& 
    \multicolumn{3}{c}{Average perturbation norm}\\
    \cmidrule(lr){2-4}   \cmidrule(lr){5-7}
    Attack & VGG19 & GoogLeNet & ResNet18 & VGG19 & GoogLeNet & ResNet18\\ 
    \midrule 
    \gls{fgsm}     & 39.97\% & 39.85\% & 40.18\% & 
    17.6232 & 0.2575 & 2.7183\\
    \gls{bim}      & 5.17\% & 4.29\% & 4.49\% & 
    8.9903 & 0.0484 & 0.2303 \\
    Boundary & 8.99\% & 25.75\% & 1.39\% & 
    0.0515 & 0.0209 & 0.0849\\
    \gls{cw}       & 4.75\% & 0.55\% & 0.30\% & 
    0.2461 &0.0140  & 0.0559 \\ 
    \bottomrule
    \end{tabular}
    \label{tab:acc}
\end{table}

\begin{figure}[h!]
	\centering
	\addtolength{\tabcolsep}{5pt}
	\begin{tabular}{rc} 
	    \gls{fgsm} & \includegraphics[height=1.2cm, align=c]{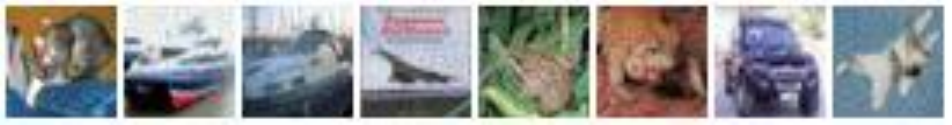}\\
	    \gls{bim} & \includegraphics[height=1.2cm, align=c]{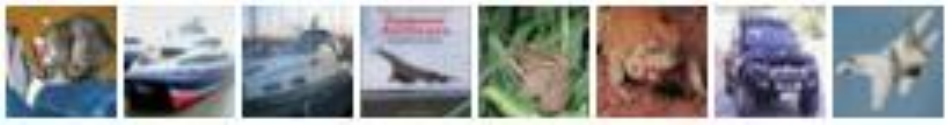}\\
	    Boundary & \includegraphics[height=1.2cm, align=c]{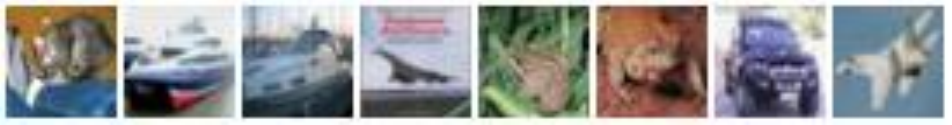}\\
	    \gls{cw} & \includegraphics[height=1.2cm, align=c]{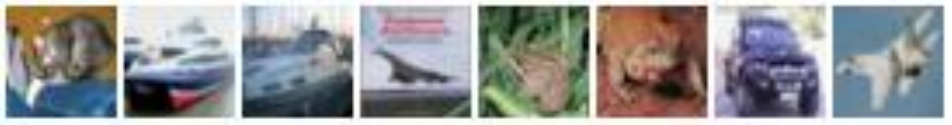}\\
	    Original & \includegraphics[height=1.2cm, align=c]{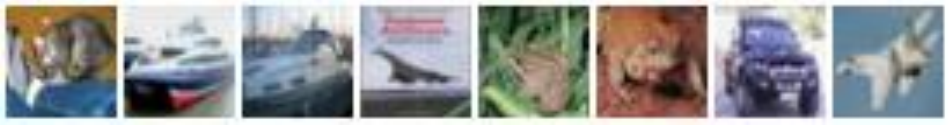}\\
	\end{tabular}
	\caption{Adversarial perturbations calculated for the ResNet18.}
	\label{fig:perturbations}
\end{figure}

In Fig.~\ref{fig:perturbations}, we visualize eight randomly selected, perturbed images from attacks on the ResNet18.
Most of the adversarial examples are unnoticeable to the human eye compared to the original image, especially those produced by the \gls{cw} and the boundary methods.
However, the perturbations calculated by the \gls{fgsm} are more recognizable since the backgrounds of the images are altered.
Similar conclusions can be made when observing the perturbed images generated from the other \glspl{dnn}, i.e., VGG19 and GoogLeNet.

\begin{table}[h!]
    \caption{Comparison of adversarial attack classification performance for single attacks.}
    \label{tab:detection}
    \centering
    \addtolength{\tabcolsep}{4pt}
    \scalebox{0.9}{%
    \begin{tabular}{llrrrr} 
    \toprule
    & & \multicolumn{2}{c}{Accuracy} & \multicolumn{2}{c}{$F_1$ score} \\
    \cmidrule(lr){3-4}  \cmidrule(lr){5-6} 
    Model & Attack & Kwon et al.~\cite{kwon2021classification} & Ours & Kwon et al.~\cite{kwon2021classification} & Ours\\
    \midrule 
    \multirow{4}{*}{VGG19} 
                           & FGSM & 71.60\% & \textbf{82.08\%} & 68.43\% & \textbf{82.05\%}\\ 
                           & BIM & 85.20\% & \textbf{98.70\%} & 84.47\% & \textbf{98.69\%}\\
                           & Boundary & \textbf{97.53\%} & 96.30\% & \textbf{97.44\%} & 96.25\%\\ 
                           & CW & 89.90\% & \textbf{90.05\%} & 89.99\% & \textbf{90.16\%}\\
    \midrule
    \multirow{4}{*}{GoogLeNet} 
                           & FGSM & 72.60\% & \textbf{76.05\%}& 73.69\% & \textbf{74.48\%}\\ 
                           & BIM & 81.50\% & \textbf{83.60\%} & 77.88\% & \textbf{82.38\%} \\
                           & Boundary & \textbf{96.50\%} & 95.50\% & \textbf{96.35\%} & 95.45\%\\ 
                           & CW & 93.65\%& \textbf{93.80\%} & 93.58\%& \textbf{93.76\%}\\
    \midrule
    \multirow{4}{*}{ResNet18} 
                           & FGSM & 70.40\% & \textbf{72.58\%} & 69.23\% & \textbf{71.37\%}\\ 
                           & BIM & 85.48\% & \textbf{89.48\%} & 83.68\% & \textbf{88.96\%}\\
                           & Boundary & \textbf{97.20\%} & 96.28\% & \textbf{97.10\%} & 96.19\%\\
                           & CW & 93.53\% & \textbf{93.58\%} & 93.63\% & \textbf{93.65\%}\\
    \bottomrule
    \end{tabular}
    }
\end{table}

Ultimately, we are interested in detecting these adversarial examples, i.e., classifying whether an input sample is adversarial or not. 
To this end, we compare the classification performance of our proposed \gls{svm}-based method to the method introduced in~\cite{kwon2021classification}.
As described in Section~\ref{sec:method}, their method only thresholds the difference between the highest and second highest class score and therefore disregards important information.
The classification performance, i.e., the binary classification accuracy and the $F_1$ score, for the various attacks on the different models are depicted in Table~\ref{tab:detection}.
We observe that our method outperforms theirs in all cases except for the boundary attack, however, the difference is small for all \glspl{dnn}.
This could be explained by the fact that the boundary attack is a black-box attack which only considers the predicted class and not the class scores.
On the other hand, we observe that our method increases the performance of detecting adversarial images produced by \gls{bim} and \gls{fgsm} by more than $10\%$ for the pre-trained VGG19.
It is also able to better detect adversarial examples generated by \gls{bim} and \gls{fgsm} on GoogLeNet and ResNet18.

\begin{table}
    \caption{Comparison of adversarial attack classification performance for combined attacks.}
    \label{tab:detection-combined}
    \centering
    \addtolength{\tabcolsep}{4pt}
    \scalebox{0.9}{%
    \begin{tabular}{llrrrr} 
    \toprule
     & & \multicolumn{2}{c}{Accuracy} & \multicolumn{2}{c}{$F_1$ Score} \\
    \cmidrule(lr){3-4}  \cmidrule(lr){5-6} 
    Model & Attack & Kwon et al.~\cite{kwon2021classification}& Ours & Kwon et al.~\cite{kwon2021classification} & Ours\\
    \midrule 
    \multirow{4}{*}{VGG19} 
                           & CW+BIM & 67.38\% & \textbf{89.90\%} & 54.80\% & \textbf{90.08\%}\\
                           & CW+FGSM & 80.75\% & \textbf{83.65\%} & 79.90\% & \textbf{83.14\%}\\
                           & Boundary+BIM & 73.45\% & \textbf{95.88\%} & 63.73\% & \textbf{95.85\%}\\
                           & Boundary+FGSM & 82.45\% & \textbf{85.80\%} & 81.92\% & \textbf{84.85\%} \\
    \midrule
    \multirow{4}{*}{GoogLeNet} 
                           & CW+BIM & 70.93\% & \textbf{84.08\%} & 59.66\% & \textbf{83.92\%}\\
                           & CW+FGSM & 79.68\% & \textbf{82.35\%} & 79.28\% & \textbf{81.37\%}\\
                           & Boundary+BIM & 73.60\% & \textbf{84.93\%} & 63.89\% & \textbf{84.57\%}\\
                           & Boundary+FGSM & 78.93\% & \textbf{80.93\%} & 78.53\% & \textbf{79.58\%}\\
    \midrule
    \multirow{4}{*}{ResNet18} 
                           & CW+BIM & 70.45\%  & \textbf{88.30\%} & 60.49\%  & \textbf{88.40\%}\\
                           & CW+FGSM & 78.68\%  & \textbf{79.33\%} & 79.03\%  & \textbf{79.52\%}\\
                           & Boundary+BIM & 72.73\%  & \textbf{90.05\%} & 62.16\%  & \textbf{89.61\%}\\
                           & Boundary+FGSM &  77.93\% & \textbf{78.85\%} &  \textbf{78.31\%} & 77.76\%\\
    \bottomrule
    \end{tabular}
    }
\end{table}
Since the defender might not know a priori which attack the attacker will use, it would be useful for an adversarial example detection method to be capable of detecting a combination of different attacks.
To this end, we simulate combinations of attacks and summarize the classification performance in Table~\ref{tab:detection-combined}.
We observe that our \gls{svm}-based method performs better in almost all cases.
In combinations of attacks which include \gls{bim}, our method improves the detection performance both in terms of the binary classification accuracy and the $F_1$ score.

\section{Conclusion}
In this paper, we propose a \gls{svm}-based method that detects adversarial attacks using the class scores of a \gls{dnn}.
Our detection method does not require one to retrain the model and achieves reasonable detection performance.
Moreover, we are able to show that our method is capable to detect adversarial images produced by different attacks at the same time.
These results indicate the potential of training adversarial example detection methods only using the class scores.
This is a simple and effective measure to help defend against adversarial attacks, which is especially important in safety-critical applications.

\bibliographystyle{splncs04}
\bibliography{bibliography}

\end{document}